\documentclass{article}

\usepackage{arxiv}
\usepackage[utf8]{inputenc}
\usepackage[T1]{fontenc}
\usepackage{amsmath,amssymb,amsfonts}
\usepackage{booktabs}
\usepackage{graphicx}
\usepackage{subcaption}
\usepackage{float}
\usepackage{microtype}
\usepackage[numbers,sort&compress]{natbib}
\usepackage{url}
\usepackage[hidelinks,bookmarksopen=true,bookmarksnumbered=true]{hyperref}

\title{Retrieval-Augmented Visual Prompting: Guiding Foundation Models in Two-Photon Imaging}

\author{
\textbf{Salvatore Calcagno \quad Marco Finocchiaro \quad Giovanni Bellitto}\\
\textbf{Daniela Giordano \quad Concetto Spampinato \quad Federica Proietto Salanitri}\\[0.6em]
\normalfont PeRCeiVe Lab\\
\normalfont Department of Electrical, Electronic and Computer Engineering\\
\normalfont University of Catania, Catania, Italy
}

\date{}
\renewcommand{\headeright}{Preprint}
\renewcommand{\undertitle}{Preprint}
\renewcommand{\shorttitle}{RAVP}

\hypersetup{
  pdftitle={Retrieval-Augmented Visual Prompting: Guiding Foundation Models in Two-Photon Imaging},
  pdfsubject={Computer Vision and Pattern Recognition; Neurons and Cognition},
  pdfauthor={Salvatore Calcagno, Marco Finocchiaro, Giovanni Bellitto, Daniela Giordano, Concetto Spampinato, Federica Proietto Salanitri},
  pdfkeywords={two-photon calcium imaging, neuron segmentation, foundation models, visual prompting, SAM 3}
}

\begin{document}
\maketitle

\begin{abstract}
Two-photon calcium imaging presents a challenging setting for foundation models: image appearance varies substantially across recordings and experimental conditions, annotations are scarce, and rapid adaptation is often needed. Rather than adapting model weights through fine-tuning, we ask whether a foundation model can be guided at inference time by injecting external visual memory directly into its input. We implement this idea with SAM~3 and introduce \emph{Retrieval-Augmented Visual Prompting} (RAVP), a framework in which each target tile is augmented with a retrieved annotated exemplar whose bounding box is used as a concept prompt.

RAVP turns retrieval into a form of visual prompting and enables adaptation through input design alone. We study multiple exemplar selection strategies, including fluorescence-guided heuristics and a lightweight recall predictor trained to estimate which exemplar is most informative for a target tile. Experiments on the Allen Brain Observatory show that exemplar-augmented inference consistently strengthens zero-shot neuron detection and instance segmentation. Ablation studies further show that a single carefully selected exemplar is more effective than prompting with multiple retrieved examples. These results position inference-time visual memory injection as a simple and effective alternative to parameter adaptation for foundation models in specialized biomedical imaging.

\end{abstract}


\section{Introduction}

\label{sec:intro}
Two-photon calcium imaging (2PCI) is a cornerstone technique in systems neuroscience, enabling \emph{in vivo} recording of neural population activity at single-cell resolution~\cite{stosiek2003,chen2013}. Automatic segmentation of neuronal regions of interest is a critical preprocessing step, as it determines which pixels are assigned to each cell and thereby conditions downstream analyses of neural dynamics. Despite decades of effort, robust and generalizable neuron segmentation remains an open challenge.

Several factors make this problem particularly difficult. In maximum intensity projections (max projections) of fluorescence videos, highly active neurons appear as bright, well-defined ellipsoidal blobs, whereas weakly active or silent neurons are nearly indistinguishable from background. In addition, recording conditions, microscope setups, cortical depth, animal-specific factors, and calcium indicators induce substantial cross-session and cross-animal domain shift. Cell density also varies markedly across brain regions, while neighboring neurons often have ambiguous or overlapping boundaries. At the same time, expert annotations are expensive and time-consuming, making large annotated corpora rare.

The advent of large-scale vision foundation models---most notably the Segment Anything Model (SAM) family~\cite{sam,sam2,sam3}---has opened new possibilities for generalizable segmentation pipelines. In particular, SAM~3~\cite{sam3} introduces \emph{Promptable Concept Segmentation} (PCS), which allows the model to detect and segment all instances of a concept from image exemplars or a text prompt in a single forward pass. This few-shot capability is especially appealing in specialized domains, where the target concept cannot be easily described in natural language but can be demonstrated visually.


However, applying SAM~3 directly to 2PCI max projections for instance segmentation remains non-trivial, as its few-shot capabilities do not reliably transfer to this domain. In particular, exemplars defined by bounding boxes within the target image face two main limitations: they require specialized manual annotation and may fail to capture the full appearance variability of the concept. Moreover, it remains unclear which specific examples are most informative for a given target image. This suggests a broader question: rather than adapting model parameters, can we adapt the \emph{visual context} presented to the model at inference time?

In this paper, we answer this question with \emph{Retrieval-Augmented Visual Prompting} (RAVP), a framework that guides SAM~3 at inference time by modifying its visual input, without changing model weights or architecture. Our key idea is a \emph{retrieval-augmented image composition} strategy: each input tile is surrounded by exemplar crops retrieved from an external annotated library, whose bounding boxes serve as concept prompts. Using an external library removes the need for target-specific annotation and provides access to a broader range of visual appearances. The library thus acts as a visual memory queried at inference time. RAVP turns exemplar retrieval and image composition into an inference-time adaptation mechanism, injecting retrieved exemplars directly into the image plane and processing them through SAM~3's native Promptable Concept Segmentation interface.

A key challenge is selecting which exemplars to retrieve. Because neuronal appearance in 2PCI spans a wide fluorescence range, not all exemplars are equally informative for a given target image. We therefore study multiple selection strategies, including fluorescence-guided retrieval heuristics, and introduce a recall predictor that estimates which exemplar is likely to maximize detection recall. We evaluate RAVP in the zero-shot setting against full fine-tuning, encoder-only fine-tuning, and LoRA~\cite{lora}. Experiments on the Allen Brain Observatory~\cite{abo} show that exemplar-augmented inference improves zero-shot performance and provides a competitive alternative to parameter adaptation.



Beyond the specific application to 2PCI, our work opens a broader design space for inference-time adaptation of foundation models in specialized biomedical imaging. The main contributions of this paper are:
\begin{enumerate}
    \item A \textbf{retrieval-augmented image composition} strategy that turns an external exemplar library into visual memory for inference-time adaptation of SAM~3, without modifying its weights or architecture and while leveraging its native exemplar-prompting interface.
    \item A systematic analysis of \textbf{exemplar selection strategies}, including fluorescence-guided adaptive selectors.
    \item A lightweight \textbf{recall predictor} $\pi_\theta$ that learns to select informative exemplars at inference time without ground-truth target annotations, trained on SAM~3 inference outcomes as a supervision signal.
    \item A comprehensive evaluation on the \textbf{Allen Brain Observatory}, showing that visual memory injection consistently strengthens zero-shot performance and provides a competitive alternative to other adaptation regimes.
\end{enumerate}

\section{Related Work}

\subsection{Neuron Segmentation in Two-Photon Calcium Imaging}
The standard pipeline for calcium imaging analysis has traditionally relied on statistical source separation. Constrained Non-negative Matrix Factorization (CNMF)~\cite{pnevmatikakis2016simultaneous} established a principled formulation for demixing spatial footprints and temporal traces from background activity by factorizing the observed fluorescence into non-negative spatial and temporal components under autoregressive calcium dynamics. While highly accurate, CNMF remains computationally demanding and operates in batch mode, where access to the full recording is assumed for reliable optimization and component refinement. 

To improve scalability, subsequent systems introduced more efficient initialization and processing strategies. \textit{Suite2P}~\cite{pachitariu2017suite2p} accelerated large-scale analysis through clustering-based initialization and optimized processing, making it suitable for high-throughput recording while still remaining essentially an offline pipeline. In the same family, \textit{OASIS}~\cite{friedrich2017fast} addressed the temporal \emph{deconvolution} problem with an online sparse non-negative formulation, enabling fast inference of spike trains from fluorescence traces. However, \textit{OASIS} does not solve spatial segmentation and must operate on traces that are already extracted. 

To enable real-time applications, online variants of dictionary learning were developed. \textit{OnACID}~\cite{giovannucci2019caiman} extended CNMF into an online framework by updating sufficient statistics frame-by-frame. This design supports simultaneous denoising, deconvolution, and source discovery, but it still depends on a warm-start initialization phase and on accumulating temporal evidence before new components can be reliably detected. 
Recently, \textit{realSEUDO}~\cite{dmitrieva2024realseudo} further refined this paradigm by introducing a real-time feedback loop that estimates activity for known neurons while continuously searching the residual for previously unseen sources. Dictionary-learning approaches remain highly effective, but they still rely on accumulating sufficient statistics over time, which can limit responsiveness in short sequences and in cold-start settings with sparse or transient neuronal activity. They are therefore conceptually distinct from promptable image-level inference with foundation models. 

\subsection{Deep Learning for Neuron Segmentation}
Deep learning has emerged as an alternative to statistical source separation pipelines by learning spatial and morphological priors directly from annotated data. \textit{CITE-On}~\cite{sita2022deep} showed that convolutional detectors can support real-time neuron localization and identity tracking from streaming frames. Unlike classical methods, this family of approaches can directly model spatial patterns at the image level, reducing reliance on explicit factorization assumptions. More general frameworks based on U-Net-like architectures have further improved instance segmentation through morphological post-processing~\cite{wu2022general}.

Self-supervised and hybrid training strategies have also been explored to mitigate the annotation bottleneck.
\textit{NeuroSeg-III}~\cite{wu2024neuroseg} combines self-supervised pre-training with an efficient detection backbone and reports strong performance on public benchmarks such as the Allen Brain Observatory~\cite{abo}.
More broadly, these approaches show that learned visual priors can be highly effective for neuron segmentation. At the same time, they remain largely frame-wise and spatially focused, typically relying on inputs such as maximum projections or correlation maps rather than explicitly addressing inference under limited temporal evidence. 
We therefore focus on a complementary direction, where a general-purpose foundation model is guided at inference time through retrieved visual exemplars instead of being specialized through additional domain-specific training.

\subsection{SAM and Its Adaptations}
The Segment Anything Model (SAM) family~\cite{sam, sam2, sam3} established a new paradigm for promptable, class-agnostic segmentation.
SAM~\cite{sam} introduced interactive segmentation from sparse geometric prompts, and SAM~2~\cite{sam2} extended this setting from images to videos through a memory-based architecture that propagates segmentation information over time, enabling promptable visual segmentation in temporally evolving scenes. More recently, SAM~3~\cite{sam3} generalized the prompting interface further through \emph{Promptable Concept Segmentation} (PCS), in which the target concept can be specified by a short noun phrase, an image exemplar, or a combination of both. In this formulation, the model is asked to detect, segment, and, in videos, track all instances matching a prompted concept, making exemplar-based prompting a native mechanism rather than an auxiliary add-on.

This flexibility has rapidly motivated domain-specific adaptations. MedSAM~\cite{medsam}, for example, adapts SAM to medical data through large-scale fine-tuning, while subsequent works have continued this direction through full-finetuning or parameter-efficient adaptation modules tailored to specialized domains. Our approach follows an orthogonal direction: instead of adapting the model weights, we adapt the \emph{input context} presented to the model, using retrieved exemplars as inference-time visual memory.

\subsection{Few-Shot Segmentation and Retrieval-Augmented Vision}
Few-shot segmentation has long studied how a small support set can define the target concept at inference time, typically through meta-learning and support-set conditioning~\cite{prototypical,panet,hsnet}. In parallel, visual in-context learning methods such as Painter~\cite{painter} and SegGPT~\cite{wang2023seggpt} showed that segmentation can emerge from support image-mask pairs presented directly at inference, without task-specific retraining. These approaches suggest that model behavior can be shaped not only by parameter updates, but also through the design of the input context.

Our work is also related in spirit to retrieval-augmented methods, where external examples or memory are used to enrich inference. The key difference is that we do not introduce a separate retrieval-and-fusion architecture. Instead, we embed retrieved exemplars directly into the image plane and exploit SAM~3's native exemplar-prompting interface as a bridge between retrieval and segmentation. This yields a simple form of retrieval-augmented visual prompting that is particularly well suited to specialized biomedical domains, where annotated support examples are scarce and full model adaptation may be impractical.


\section{Method}
\label{sec:method}

\subsection{Problem Formulation}
\label{sec:formulation}

Let $I \in \mathbb{R}^{H \times W}$ be a single-channel 2PCI max projection image. Our goal is to predict a set of neuronal ROI detections
\begin{equation}
    \mathcal{D}(I) = \bigl\{ (b_i,\, m_i,\, s_i) \bigr\}_{i=1}^{N},
    \label{eq:detections}
\end{equation}
where $b_i \in \mathbb{R}^4$ is a bounding box in $(x, y, w, h)$ format,  $m_i \in \{0,1\}^{H \times W}$ is a binary segmentation mask, and $s_i \in [0,1]$ is a confidence score.

We operate in a few-annotation regime, and assume access to an annotated support \emph{library} $\mathcal{L} = \{(c_j,\, \mathcal{M}_j,\, f_j)\}_{j=1}^{|\mathcal{L}|}$, where $c_j \in \mathbb{R}^{H_c \times W_c}$ is an image crop 
centered on one annotated neuron, $\mathcal{M}_j$ is the associated set of ground-truth masks, and $f_j = \mathrm{mean}(\{s : s \in c_j,\, s \in \text{neuron}\})$ is the mean fluorescence of annotated neurons in $c_j$. The library is built exclusively from a held-out split, disjoint from 
training and evaluation data.

\begin{figure*}[ht!]
  \centering
  \includegraphics[width=\linewidth, trim={0 5cm 0 0}, clip]{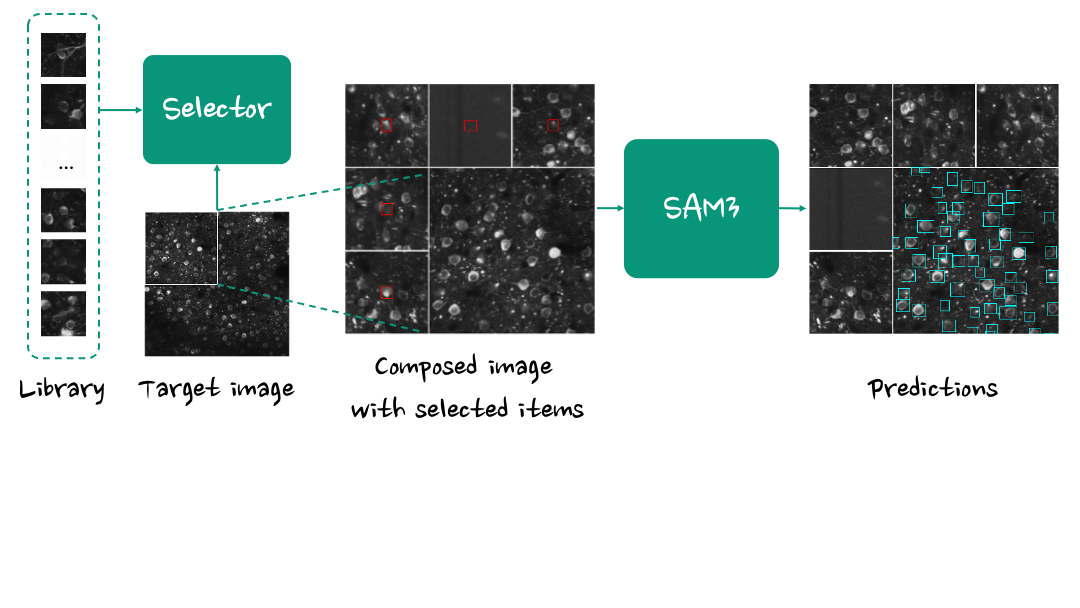}
  \caption{Overview of the proposed RAVP framework. A Selector retrieves exemplar crops from an external annotated Library. These crops are tiled along the borders of the Target image tile to create a Composed image. This composite is passed as a single input to SAM~3; the bounding boxes of the annotated exemplars serve as visual prompts for the model's Promptable Concept Segmentation head to generate the final Predictions across the target region.}
\end{figure*}

\subsection{Retrieval-Augmented Image Composition}
\label{sec:library}

The central contribution of this work is a mechanism for injecting visual memory directly into the input of SAM~3, without modifying model weights or architecture. For each target tile $T_p$\footnote{As justified in Section \ref{sec:implementation}, we process the target image using a sliding window approach rather than analyzing it in its entirety. This is because SAM~3 has a constrained prediction capacity that cannot simultaneously account for all neurons; we therefore partition the image into individual tiles $T_p$.}, we construct a composite image $\tilde{I}_p$ by augmenting $T_p$ with a border of exemplar crops retrieved from $\mathcal{L}$. Formally, let $\{c_{j_1}, \ldots, c_{j_k}\}$ be the selected exemplars and let $\phi_p$, $\phi_{j_\ell}$ denote the affine maps from tile and exemplar coordinates to canvas coordinates, respectively. The composite image $\tilde{I}_p$ is constructed by placing $T_p$ at the center of a canvas $\mathcal{C}$ and tiling the retrieved crops along top and left sides. 
All ground-truth annotations are projected onto the canvas coordinate system via $\tilde{b} = \phi(b)$, $\tilde{m} = \phi(m)$.

The composite image $\tilde{I}_p$ is passed to SAM~3 as a single input. The projected bounding boxes of annotated neurons in the border region serve as concept prompts $\mathcal{P} = \{\tilde{b}_{j_\ell, i}\}$ for SAM~3's Promptable Concept Segmentation head, which detects and segments all matching instances across the full canvas. Predictions are then restricted to the target region via the inverse map $\phi_p^{-1}$.

At inference time, ground-truth annotations from the target tile $T_p$ are used exclusively for quantitative evaluation; only the bounding boxes annotated on the retrieved library crops are provided as visual prompts to guide the detection process. Each exemplar is extracted with a surrounding context margin $\delta > 0$: a neuron with ground-truth box $b = (x, y, w, h)$ yields a crop at $(x - \delta,\, y - \delta,\, w + 2\delta,\, h + 2\delta)$. This provides SAM~3 with sufficient spatial context to form a reliable concept representation, and avoids boundary ambiguities that arise from overly tight crops.

\subsection{Exemplar Selection Strategies}
\label{sec:selection}

A critical challenge in 2PCI is the extreme variability of neuronal fluorescence, which manifests both within and across imaging sessions. Locally, highly active neurons appear as bright, well-defined blobs, while inactive ones are nearly invisible against the background. Globally, different target images exhibit significant variations in background noise, labeling density, and overall signal-to-noise ratio due to diverse experimental conditions. Consequently, randomly chosen exemplars are unlikely to capture the specific appearance range of a given target image, leading to suboptimal prompting. We therefore study the following selection strategies:

\paragraph{Random.}
We sample \(k\) items randomly from the library $\mathcal{L}$:
\begin{equation}
    \mathcal{J}(T_p) \sim \mathrm{Uniform}\bigl(\{\mathcal{J} \subseteq \{1,\dots,|\mathcal{L}|\}: |\mathcal{J}|=k\}\bigr).
\end{equation}
\paragraph{Fixed fluorescence.}
We rank library items according to the distance between their mean fluorescence and a fixed target value \(\bar{f}\), and select the \(k\) closest exemplars:
\begin{equation}
    \mathcal{J}(T_p) = \operatorname{TopK}_{j}\bigl(-|f_j-\bar{f}|\bigr).
    \label{eq:fixed}
\end{equation}

\paragraph{Image-adaptive.}
We adapt exemplar selection to the current tile using its mean intensity as a simple proxy for local imaging conditions. Let \(\hat{f}(T_p)\) denote the mean fluorescence of \(T_p\), computed over all tile pixels. We then select the \(k\) library items whose fluorescence is closest to \(\hat{f}(T_p)\):
\begin{equation}
    \mathcal{J}(T_p) = \operatorname{TopK}_{j}\bigl(-|f_j-\hat{f}(T_p)|\bigr).
    \label{eq:adaptive}
\end{equation}
These strategies allow us to study whether matching exemplars to the fluorescence regime of the target image improves concept prompting in SAM~3.

\subsection{Recall Predictor for Learned Exemplar Selection}
\label{sec:predictor}

The heuristic strategies above rely on image statistics as a proxy for exemplar quality. To move beyond hand-crafted solutions, we introduce a lightweight \emph{recall predictor} $\pi_\theta$, which estimates how effective a candidate exemplar will be for a given target tile. 

For a tile--exemplar pair \((T_p, c_j)\), the predictor outputs: 
\begin{equation}
    \hat{r}(T_p, c_j) = \pi_\theta(T_p,\, c_j).
    \label{eq:predictor}
\end{equation}
where \(\hat{r}(T_p, c_j)\) is the predicted recall that SAM~3 would achieve on \(T_p\) when prompted with \(c_j\).
At inference time, the optimal exemplar is selected as:
\begin{equation}
    j^* = \arg\max_{j \in \mathcal{L}} \pi_\theta(T_p,\, c_j).
    \label{eq:pred_selection}
\end{equation}

The predictor is trained offline using supervision derived from actual SAM~3 inference outcomes on the training split. For each pair \((T_p, c_j)\), we run SAM~3 using \(c_j\) as support exemplar, compute the achieved recall \(r(T_p, c_j)\), and train \(\pi_{\theta}\) by minimizing
\begin{equation}
    \mathcal{L}_{\pi} =
    \frac{1}{|\mathcal{Q}_{\mathrm{train}}|}
    \sum_{(T_p, c_j) \in \mathcal{Q}_{\mathrm{train}}}
    \bigl(\pi_{\theta}(T_p, c_j) - r(T_p, c_j)\bigr)^2,
    \label{eq:pred_loss}
\end{equation}
where \(\mathcal{Q}_{\mathrm{train}}\) denotes the set of training tile--exemplar pairs. Because \(\pi_{\theta}\) is lightweight, it can score a large number of candidate exemplars before a single SAM~3 forward pass. Architectural details of the predictor are provided in Section~\ref{sec:implementation}.

\section{Experimental Results}
\label{sec:experiments}

\subsection{Dataset}
\label{sec:datasets}

We use the \emph{Allen Brain Observatory} (ABO)~\cite{abo}, a large-scale public repository of in vivo two-photon calcium imaging recordings from mouse visual cortex, spanning multiple cortical areas (VISp, VISl, VISam, VISal, VISpm, VISrl), imaging depths, and transgenic lines expressing calcium indicators. Raw fluorescence videos are converted to maximum intensity projections over time, yielding single-channel grayscale images that capture the spatial footprint of all active neurons across the recording session. Ground-truth ROI annotations provide per-instance binary segmentation masks. The dataset presents substantial intra- and inter-session variability in neuronal density, fluorescence range, and background structure, making it a challenging benchmark for generalizable segmentation.

To ensure a rigorous evaluation of the model's generalization capabilities, the data split is performed at the subject (animal) level. This ensures that no imaging sessions from any animal used for training or library construction appear in the test sets, thereby preventing data leakage related to individual anatomical features. Specifically, the data are partitioned as follows:
\begin{itemize}
    \item \emph{Library}: held-out 96 sessions used exclusively to construct the exemplar library $\mathcal{L}$.
    \item \emph{Train}: 462 sessions used for fine-tuning all trainable components, including the SAM~3 adaptation strategies and the recall predictor $\pi_\theta$.
    \item \emph{Validation}: 66 sessions drawn from the same imaging depths as train, used exclusively for model selection, hyperparameter tuning, and design choices such as post-processing thresholds, exemplar layout, and library size.
    \item \emph{Test (ABO in-domain)}: 73 ABO sessions from experimental conditions similar to training, annotated with the original ABO ground truth.
    \item \emph{Test (STNeuroNet subset)}: 20 additional ABO sessions for which independent ROI annotations are available from STNeuroNet~\cite{stneuronet}. This split keeps imaging conditions identical to the ABO in-domain test data, but replaces the ground truth with STNeuroNet masks.
    \item \emph{Test (OOD)}: 17 ABO sessions imaging depths never seen during training, annotated with the original ABO ground truth.
\end{itemize}

\subsection{Adaptation Strategies and Baselines}
\label{sec:baselines}

We evaluate SAM~3~\cite{sam3}, the natural reference model for RAVP, as it introduces Promptable Concept Segmentation, where a visual exemplar defines a concept and the model segments all matching instances. Earlier SAM variants lack this interface, while domain-specific neuron segmentation methods rely on different input or supervision assumptions. 

As baselines for parameter adaptation, we first evaluate SAM~3 zero-shot, using only the text prompt ``neuron'' and no parameter updates. We then compare three train-time adaptation regimes: full fine-tuning, encoder-only fine-tuning, and LoRA fine-tuning~\cite{lora}.

We evaluate Retrieval-Augmented Image Composition with a frozen, pretrained SAM~3, comparing random, fixed-fluorescence, image-adaptive, and learned exemplar selection strategies (Sections~\ref{sec:selection} and~\ref{sec:predictor}). Quantitative results are reported in Section~\ref{sec:res_library}. We also test MedSAM~3~\cite{medisam3} as an additional baseline; results are reported in the supplementary material.

\subsection{Implementation Details}
\label{sec:implementation}

\paragraph{Sliding window inference.} SAM~3 produces a bounded number of predictions per forward pass, insufficient for 2PCI images containing several hundred neuronal ROIs. To address this, we employ a sliding window approach: each image is partitioned into tiles using a 50\% overlap strategy, where each tile corresponds to half the spatial dimensions (50\% of width and height) of the original image. Each tile is processed independently, and the resulting detections are then merged into full-resolution coordinates using non-maximum suppression (NMS) to resolve redundancies in the overlapping regions.
\paragraph{Post-processing.} All predictions are passed through a two-stage pipeline before evaluation, applied uniformly across all conditions. First, detections whose bounding box area falls outside $[A_\mathrm{min}, A_\mathrm{max}] = [100, 1000]\,\mathrm{px}^2$ are removed. Second, NMS with $\tau_\mathrm{NMS} = 0.4$ suppresses duplicate detections. Both thresholds were selected empirically on a held-out validation set. Area filtering is applied before NMS to reduce the candidate set prior to the more computationally expensive suppression step.
\paragraph{Exemplar configuration.} Based on an empirical analysis of exemplar count and layout, we adopt a configuration with a single exemplar crop per tile. The exemplar is placed at a random border location (top or left) of the composite image, while the remaining border regions are filled with black padding. This design is used in all retrieval-augmented experiments and is supported by an ablation over the number of exemplars (Section~\ref{sec:res_ablation}).
\paragraph{Recall predictor.} The recall predictor $\pi_\theta$ operates on visual features rather than raw images. Given a target tile $T_p$ and a candidate exemplar $c_j$, we extract feature vectors using a frozen ResNet-18 backbone~\cite{resnet}, applied independently to each input. The two feature vectors are concatenated and passed through a two-layer MLP with ReLU activations, which regresses the predicted recall score $\hat{r}(T_p, c_j)$. The model is trained with the mean squared error loss in Eq.~\eqref{eq:pred_loss}, using recall values computed from SAM~3 outputs on the training split as supervision. Once trained, $\pi_\theta$ is used only at inference time to score library items for a given tile and select the highest-scoring exemplar.
The library $\mathcal{L}$ contains a total of 7,889 items. To maintain inference efficiency, we avoid evaluating $\pi_\theta$ for the entire library. Instead, we first filter for exemplars with sufficient spatial context and then select the first $N=100$ candidate items according to their sorted fluorescence values. The recall predictor is then queried only for this subset. We discuss the impact of this pre-selection and provide an ablation on $N$ in Section~\ref{sec:res_ablation}.
\paragraph{Training details.} 

All fine-tuning strategies (full, encoder-only and LoRA adapter) use a composite loss
\begin{equation}
    \mathcal{L} = \lambda_1 \mathcal{L}_\mathrm{box} + \lambda_2 \mathcal{L}_\mathrm{cls} + \lambda_3 \mathcal{L}_\mathrm{mask} + \lambda_4 \mathcal{L}_\mathrm{seg},
\end{equation}
where $\mathcal{L}_\mathrm{box} = \mathcal{L}_\mathrm{L1} + \mathcal{L}_\mathrm{GIoU}$, $\mathcal{L}_\mathrm{cls}$ combines focal and binary cross-entropy losses, $\mathcal{L}_\mathrm{mask}$ combines focal and Dice losses, and $\mathcal{L}_\mathrm{seg}$ is an auxiliary semantic segmentation loss. 

Fine-tuning is performed for 20 epochs with a batch size 1, using the AdamW optimizer with a weight decay of $0.1$. We adopt an inverse square root learning rate schedule with a $20$-step linear warm-up. The base learning rate is $\eta = 8 \times 10^{-5}$ for the transformer and $2.5 \times 10^{-5}$ for the vision backbone, with a layer-wise decay of $0.9$. Images are resized to $1008 \times 1008$ pixels with standard augmentations (random resizing, cropping, and intensity jitter). 
LoRA fine-tuning uses rank $r=8$, scaling factor $\alpha=16$, and dropout rate $p=0.05$ on attention modules; only the low-rank parameters are updated, while all base weights remain frozen. 

For the fixed-fluorescence selector, we set the target value to $\bar{f} = 0$, which corresponds to consistently selecting the lowest-fluorescence exemplars in $\mathcal{L}$. This choice is motivated by empirical evidence suggesting that selecting high-fluorescence neurons tends to degrade overall performance, as the Segment Anything Model (SAM) exhibits a bias toward identifying only the most prominent, high-signal structures. Conversely, low-fluorescence exemplars act as a more conservative and less restrictive prior; by avoiding over-reliance on overly salient samples, this approach ensures the model maintains sensitivity across a broader range of neuronal intensities.

All experiments are run on NVIDIA A6000 GPUs.

\section{Results}
\label{sec:results}

We report Average Precision (AP$_{50}$), Average Recall (AR$_{50}$) at IoU threshold 0.5 for both bounding box detection and instance segmentation, computed independently for each evaluation split. To provide a single balanced metric of the models' performance, we also report the harmonic mean of AP$_{50}$ and AR$_{50}$ values as H.

\subsection{Adaptation Strategies}
\label{sec:res_baseline}

Tables~\ref{tab:baseline_bbox} and ~\ref{tab:baseline_segm} summarize detection and segmentation performance across ABO evaluation splits. While the following results refer specifically to detection metrics, the same observations and performance trends extend to the segmentation task. On the ABO in-domain test set, zero-shot SAM~3 attains modest precision (AP$_{50}=0.28$) but relatively high recall (AR$_{50}=0.77$), reflecting a tendency to produce many low-quality detections on a modality it has never seen. Fine-tuning SAM~3 closes most of the gap to a high-quality detector: encoder-only and LoRA-based adaptation reach AP$_{50}=0.69$ and $0.68$, respectively, both outperforming full fine-tuning (AP$_{50}=0.64$) on the in-domain split.

On the OOD split, all fine-tuned models suffer a substantial precision drop compared to in-domain (e.g., encoder-only from AP$_{50}=0.69$ to $0.63$), but still remain far above the zero-shot baseline (AP$_{50}=0.12$), underscoring both the severity of cross-session domain shift and the benefit of any form of adaptation. A similar ranking holds on the STNeuroNet subset, where encoder-only and LoRA fine-tuning consistently dominate full fine-tuning (AP$_{50}$ up to 0.75 vs.\ 0.69), suggesting that constraining the adaptation either to the vision backbone or to low-rank adapters is preferable to unconstrained updates.

\begin{table*}[ht]
\centering
\caption{Detection performance across ABO evaluation splits. 
}
\label{tab:baseline_bbox}
\setlength{\tabcolsep}{5pt}
\resizebox{\textwidth}{!}{%
\begin{tabular}{lccccccccc}
\toprule
& \multicolumn{3}{c}{\textbf{ABO in-domain}} & \multicolumn{3}{c}{\textbf{OOD}} & \multicolumn{3}{c}{\textbf{STNeuroNet subset}}\\
\cmidrule(lr){2-4}\cmidrule(lr){5-7}\cmidrule(lr){8-10}
\textbf{Method} & AP$_{50}$ & AR$_{50}$ & H & AP$_{50}$ & AR$_{50}$ & H& AP$_{50}$ & AR$_{50}$ & H\\
\midrule
SAM~3 (zero-shot)      & 0.277 & 0.767 & 0.407 & 0.117 & 0.802 & 0.204 & 0.405 & 0.736 & 0.523 \\
\midrule
SAM~3 + Enc.\ FT       & 0.687 & 0.862 & 0.764 & 0.630 & 0.894 & 0.739 & 0.737 & 0.828 & 0.780 \\
SAM~3 + LoRA           & 0.676 & 0.850 & 0.730 & 0.646 & 0.870 & 0.741 & 0.751 & 0.844 & 0.795 \\
SAM~3 + Full FT        & 0.630 & 0.818 & 0.712 & 0.594 & 0.880 & 0.709 & 0.685 & 0.796 & 0.737 \\
\midrule
SAM~3 + Lib. \textit{(Recall Predictor)}       & 0.528 & 0.736 & 0.615 & 0.468 & 0.786 & 0.587 & 0.609 & 0.732 & 0.665 \\
\bottomrule
\end{tabular}
}
\end{table*}

\begin{table*}[ht]
\centering
\caption{Segmentation performance across ABO evaluation splits. 
}
\label{tab:baseline_segm}
\setlength{\tabcolsep}{5pt}
\resizebox{\textwidth}{!}{%
\begin{tabular}{lccccccccc}
\toprule
& \multicolumn{3}{c}{\textbf{ABO in-domain}} & \multicolumn{3}{c}{\textbf{OOD}} & \multicolumn{3}{c}{\textbf{STNeuroNet subset}}\\
\cmidrule(lr){2-4}\cmidrule(lr){5-7}\cmidrule(lr){8-10}
\textbf{Method} & AP$_{50}$ & AR$_{50}$ & H & AP$_{50}$ & AR$_{50}$ & H& AP$_{50}$ & AR$_{50}$ & H\\
\midrule
SAM~3 (zero-shot)      & 0.279 & 0.750 & 0.407 & 0.117 & 0.789 & 0.200 & 0.403 & 0.735 & 0.521 \\
\midrule
SAM~3 + Enc.\ FT       & 0.704 & 0.866 & 0.777 & 0.606 & 0.868 & 0.713 & 0.744 & 0.825 & 0.782\\
SAM~3 + LoRA           & 0.703 & 0.865 & 0.776 & 0.644 & 0.858 & 0.736 & 0.764 & 0.844 & 0.802\\
SAM~3 + Full FT        & 0.657 & 0.834 & 0.735 & 0.583 & 0.850 & 0.692 & 0.697 & 0.793 & 0.742\\
\midrule
SAM~3 + Lib. \textit{(Recall Predictor)}        & 0.541 & 0.712 & 0.615 & 0.468 & 0.732 & 0.571 & 0.613 & 0.714 & 0.660 \\
\bottomrule
\end{tabular}
}
\end{table*}

\subsection{Effect of the Library Mechanism}
\label{sec:res_library}

Table~\ref{tab:library} summarizes the effect of the Library mechanism in the zero-shot regime across all evaluation splits. On the ABO in-domain test set, zero-shot SAM~3 reaches AP$_{50}=0.28$ and H $=0.41$, whereas exemplar-based prompting consistently improves performance: even random selection increases AP$_{50}$ to 0.45 and H to 0.53, while the Recall Predictor reaches AP$_{50}=0.53$ and H $=0.62$. A similar trend is observed on the STNeuroNet subset, where the Recall Predictor raises AP$_{50}$ from 0.41 to 0.60 and H from 0.52 to 0.66, showing that a single informative exemplar can substantially improve zero-shot performance without modifying SAM~3's weights.


On the OOD split, where zero-shot SAM~3 exhibits very low precision (AP$_{50}=0.12$, H $=0.20$), all Library variants substantially improve detection quality, with H scores around $0.58$--$0.60$. The similar performance across selection strategies suggests that, under domain shift---such as changes in imaging depth---the main benefit comes from retrieval-augmented prompting itself. Valid library exemplars appear to provide a strong visual prior, making the specific selection policy less critical, as also suggested by the strong performance of random sampling.
Therefore, the OOD results primarily demonstrate the robustness of the retrieval-augmented prompting mechanism, rather than the universal superiority of the learned selector.


Compared with encoder-only and LoRA fine-tuning, these improvements are obtained at a lower adaptation cost (including the one-time offline generation of recall supervision with SAM~3,  feature extraction and recall-predictor optimization): the total cost is only $42\%$ of full fine-tuning and $55\%$ of LoRA adaptation.  Once the predictor is trained, SAM~3 remains fully frozen and requires no further parameter updates. Retrieval-augmented prompting therefore provides a cheaper adaptation path while recovering a large fraction of the performance gains of full fine-tuning.

\begin{figure*}[t]
    \centering
    \begin{subfigure}[b]{0.32\linewidth}
        \centering
        \includegraphics[width=\linewidth]{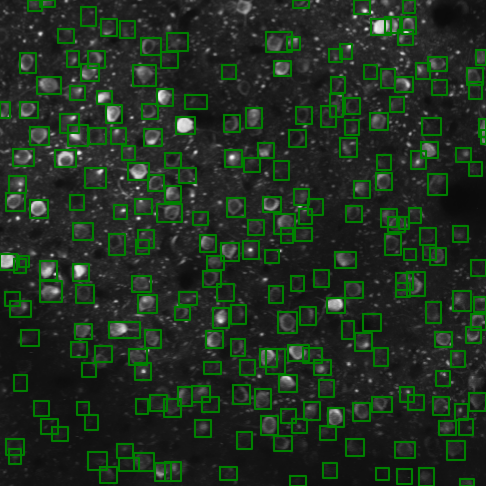}
        \caption{Ground Truth}
    \end{subfigure}
    \hfill
    \begin{subfigure}[b]{0.32\linewidth}
        \centering
        \includegraphics[width=\linewidth]{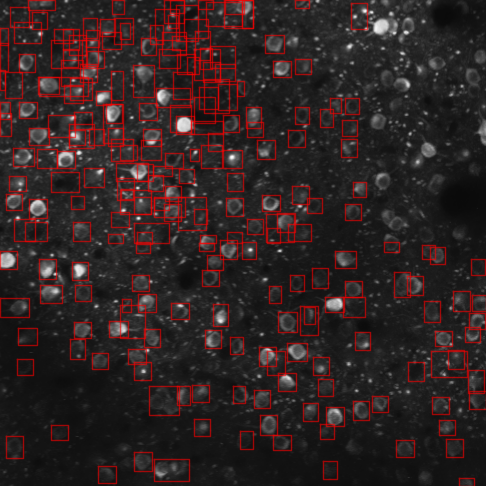}
        \caption{SAM~3 zero-shot}
    \end{subfigure}
    \hfill
    \begin{subfigure}[b]{0.32\linewidth}
        \centering
        \includegraphics[width=\linewidth]{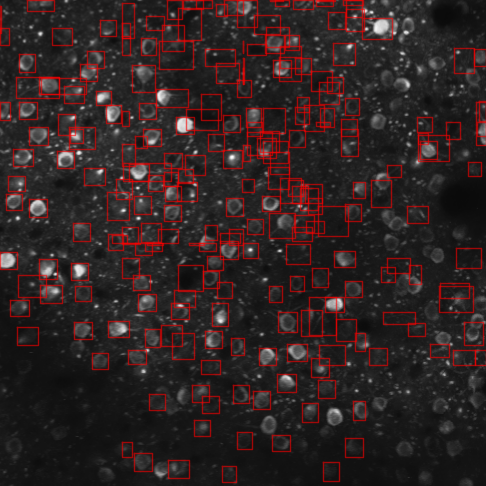}
        \caption{Random Selector}
    \end{subfigure}

    \vspace{0.3cm} 

    \begin{subfigure}[b]{0.32\linewidth}
        \centering
        \includegraphics[width=\linewidth]{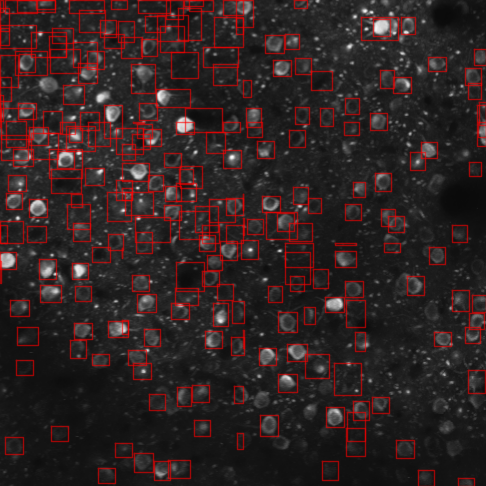}
        \caption{Image Adaptive Selector}
    \end{subfigure}
    \hfill
    \begin{subfigure}[b]{0.32\linewidth}
        \centering
        \includegraphics[width=\linewidth]{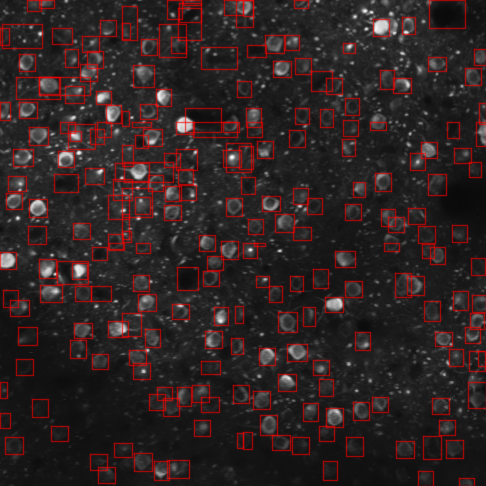}
        \caption{Fixed Fluorescence Selector}
    \end{subfigure}
    \hfill
    \begin{subfigure}[b]{0.32\linewidth}
        \centering
        \includegraphics[width=\linewidth]{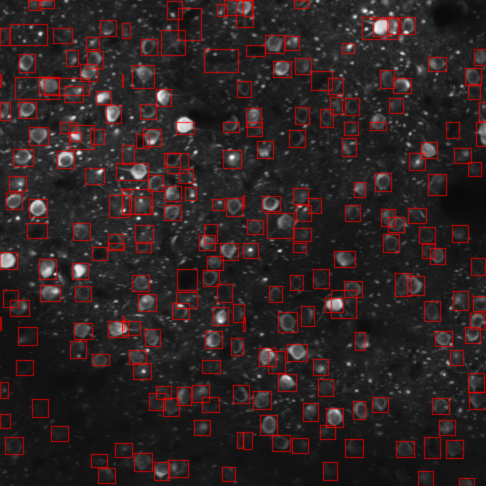}
        \caption{Recall Predictor Selector}
    \end{subfigure}

    \caption{Qualitative comparison of detection results on a representative test sample. (a) Ground truth annotations; (b) SAM~3 zero-shot performance without exemplar prompting; (c-f) Results obtained using different exemplar selection strategies. }
    \label{fig:duck_grid}
\end{figure*}

The qualitative impact of different selection strategies is illustrated in Figure \ref{fig:duck_grid}. As observed, the zero-shot baseline (Fig. \ref{fig:duck_grid}b) fails to capture the majority of neuronal structures due to the lack of domain-specific prompts. Among the proposed methods, the Fixed Fluorescence Selector and the Recall Predictor Selector (Fig. \ref{fig:duck_grid}e-f) provide the most comprehensive detections. 

\begin{table*}[t]
\centering
\caption{Effect of the Library mechanism in the zero-shot regime. AP$_{50}$ / AR$_{50}$ and H for bounding-box detection across ABO in-domain, OOD, and STNeuroNet subset. Best results per split in \textbf{bold}. For the Random Selector, averages over 10 seeds are reported.}
\label{tab:library}
\setlength{\tabcolsep}{4pt}
\resizebox{\textwidth}{!}{%
\begin{tabular}{lccc|ccc|ccc}
\toprule
& \multicolumn{3}{c}{\textbf{ABO in-domain}} & \multicolumn{3}{c}{\textbf{OOD}} & \multicolumn{3}{c}{\textbf{STNeuroNet subset}}\\
\cmidrule(lr){2-4}\cmidrule(lr){5-7}\cmidrule(lr){8-10}
\textbf{Method} & AP$_{50}$ & AR$_{50}$ & H &
                  AP$_{50}$ & AR$_{50}$ & H &
                  AP$_{50}$ & AR$_{50}$ & H \\
\midrule
SAM~3 (zero-shot)                                & 0.277 & \textbf{0.767} & 0.407 & 0.117 & \textbf{0.802} & 0.204 & 0.405 & \textbf{0.736} & 0.523 \\
\quad + Lib.\ (\textit{Random})                  & 0.446 & 0.669 & 0.535 & \textbf{0.500} & 0.752 & \textbf{0.601} & 0.473 & 0.641 & 0.544 \\
\quad + Lib.\ (\textit{Image-adaptive})          & 0.453 & 0.678 & 0.543 & 0.476 & 0.736 & 0.578 & 0.499 & 0.657 & 0.567 \\
\quad + Lib.\ (\textit{Fixed fluor.}, $f{=}0$)   & 0.469 & 0.673 & 0.553 & 0.483 & 0.751 & 0.588 & 0.565 & 0.694 & 0.623 \\
\quad + Lib.\ (\textit{Recall Predictor})        & \textbf{0.528} & 0.736 & \textbf{0.615} & 0.468 & 0.786 & 0.587 & \textbf{0.609} & 0.732 & \textbf{0.665} \\
\bottomrule
\end{tabular}
}
\end{table*}

\section{Ablation Study}
\label{sec:res_ablation}

\begin{figure*}[htb!]
  \centering
  \begin{minipage}[t]{0.4\textwidth}
    \centering
    \includegraphics[width=\linewidth]{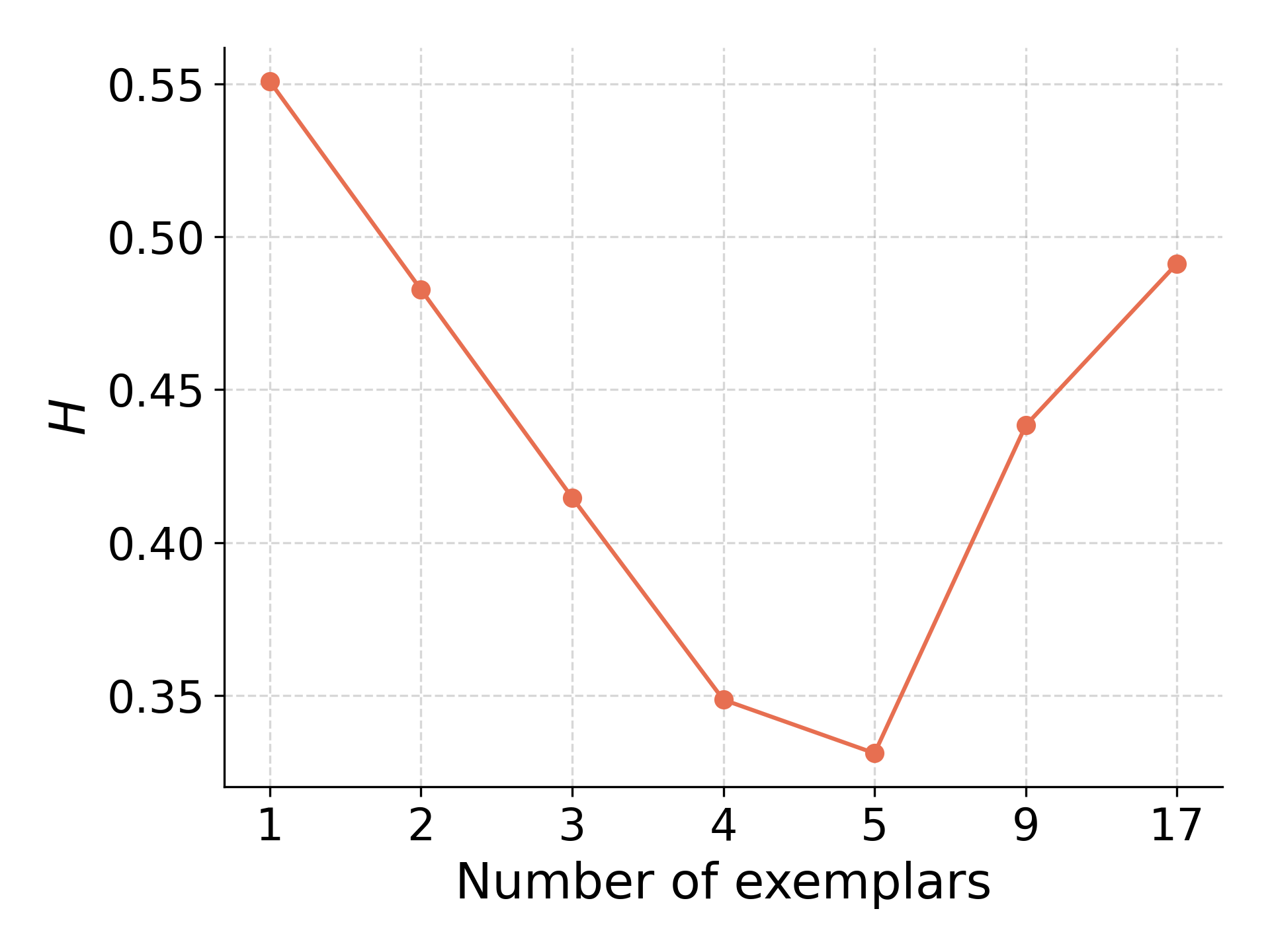}
    \caption*{(a) Number of exemplars per tile}
  \end{minipage}\hfill
  \begin{minipage}[t]{0.4\textwidth}
    \centering
    \includegraphics[width=\linewidth]{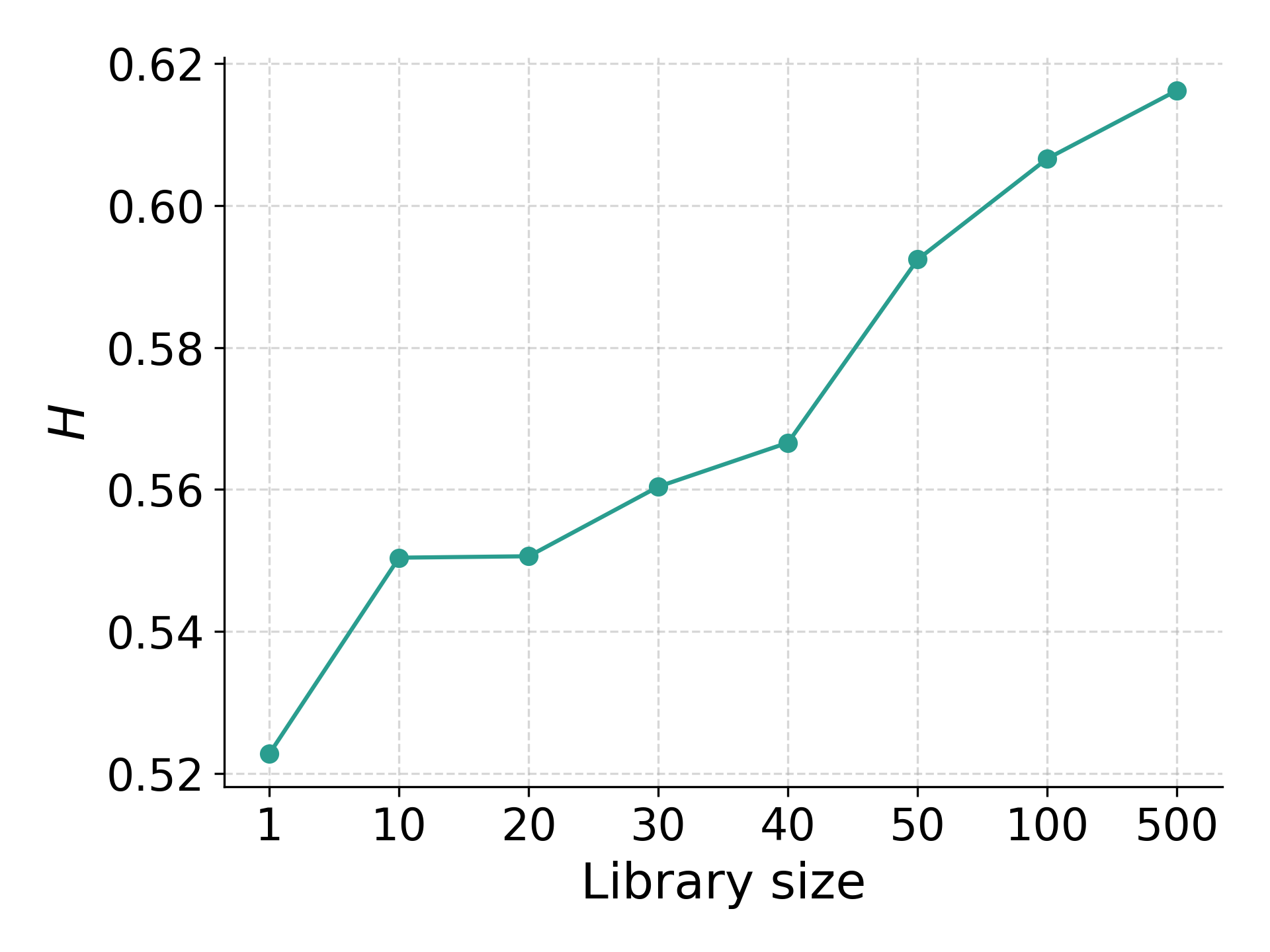}
    \caption*{(b) Library size}
  \end{minipage}
  \caption{Ablation studies on the validation split. Detection H for Library SAM as a function of (a) the number of exemplars per tile, and (b) the size of the exemplar library.}
  \label{fig:ablations}
\end{figure*}

We conduct a series of ablation studies on the validation split to understand the sensitivity of Library SAM to design choices in the construction and use of the exemplar pool (Figure~\ref{fig:ablations}). In all experiments, SAM~3 is kept frozen and only the configuration of the Library mechanism is varied. Additional controls on the canvas layout, padding, or arbitrary bounding-box cues are reported in the supplementary material.

\paragraph{Number of exemplars per tile.}
Figure~\ref{fig:ablations}(a) shows detection H as a function of the number of exemplars $k$ per tile. With a single exemplar ($k=1$), using the library attains H $\approx 0.55$. Increasing $k$ to $2,3,4,5$ leads to a monotonic degradation, with H dropping to approximately $0.33$ at $k=5$. Configurations with more exemplars partially recover recall but still underperform the single-exemplar setting. These results indicate that, in 2PCI, a single carefully selected exemplar with sufficient spatial context is more effective than presenting multiple exemplars drawn from different fluorescence regimes, and justify our choice of $k=1$ in all main experiments.

\paragraph{Library size.}
Figure~\ref{fig:ablations}(b) shows the effect of exemplar library size while using a single exemplar per tile. Increasing the library from $1$ to $50$ candidates yields most of the gain (H $\approx 0.52$ to $0.59$), whereas larger pools provide only marginal improvements. This indicates that a moderately sized library captures most of the relevant appearance variability. We therefore use $100$ candidates in all experiments as a trade-off between performance and computational efficiency.

\section{Conclusion}
\label{sec:conclusion}

We have presented Retrieval-Augmented Visual Prompting (RAVP), a framework for neuron detection and segmentation in two-photon calcium imaging that adapts SAM~3 entirely at inference time, through image-plane exemplar composition, without updating model weights or architecture. Our experiments show that a single, carefully selected exemplar can substantially improve zero-shot detection, recovering much of the gap to fine-tuned models. Adaptive selection generally outperforms random or fixed choices, while ablations show that a single exemplar from a moderately sized library captures most of the available gains, outperforming prompting with multiple retrieved examples and underscoring the importance of exemplar quality over quantity.

A limitation of the strongest variant (i.e., \textit{Recall Predictor}) is that, beyond annotating the exemplar library, it requires SAM~3 inference to generate recall supervision and training the recall predictor. Although cheaper than parameter-based adaptation, this cost remains non-negligible. Simpler selectors avoid this stage while remaining competitive, making RAVP a flexible alternative to weight-based adaptation in specialized biomedical imaging, with potential applicability beyond 2PCI to other domains where foundation models struggle to generalize.

\clearpage
\appendix
\setcounter{figure}{0}
\renewcommand{\thefigure}{S\arabic{figure}}
\setcounter{table}{0}
\renewcommand{\thetable}{S\arabic{table}}
\section{Additional Results: MedSAM~3}

\begin{figure}[H]
    \centering
    \begin{subfigure}[b]{0.32\linewidth}
        \centering
        \includegraphics[width=\linewidth]{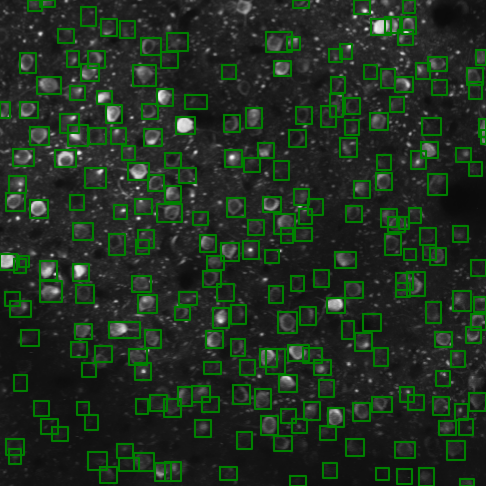}
        \caption{Ground Truth}
    \end{subfigure}
    \hfill
    \begin{subfigure}[b]{0.32\linewidth}
        \centering
        \includegraphics[width=\linewidth]{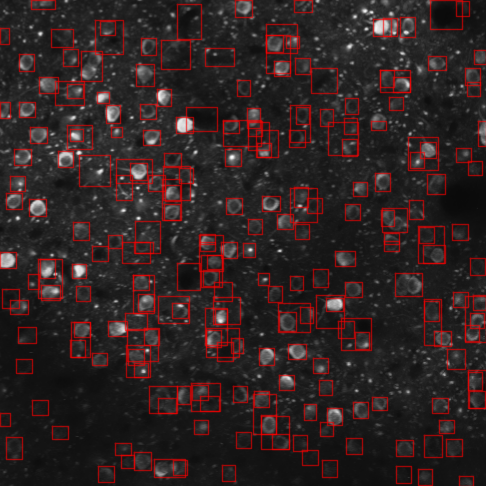}
        \caption{MedSAM~3 zero-shot}
    \end{subfigure}
    \hfill
    \begin{subfigure}[b]{0.32\linewidth}
        \centering
        \includegraphics[width=\linewidth]{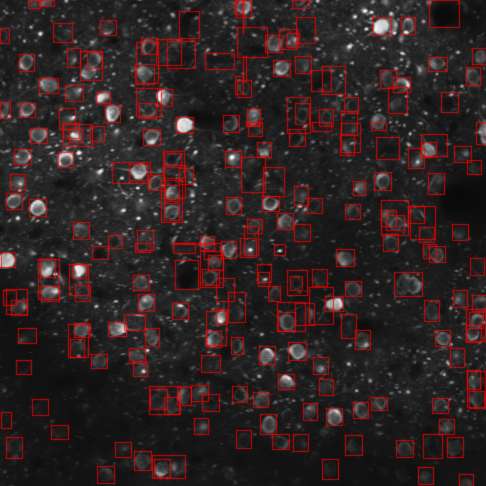}
        \caption{Random Selector}
    \end{subfigure}

    \vspace{0.3cm} 

    \begin{subfigure}[b]{0.32\linewidth}
        \centering
        \includegraphics[width=\linewidth]{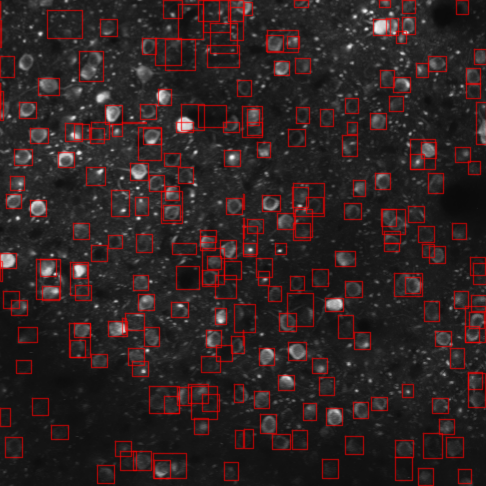}
        \caption{Image Adaptive Selector}
    \end{subfigure}
    \hfill
    \begin{subfigure}[b]{0.32\linewidth}
        \centering
        \includegraphics[width=\linewidth]{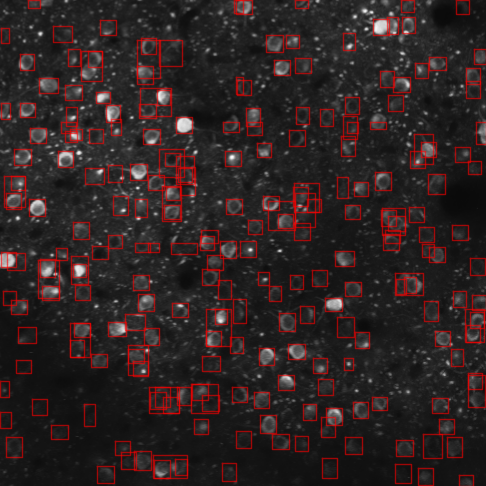}
        \caption{Fixed Fluorescence Selector}
    \end{subfigure}
    \hfill
    \begin{subfigure}[b]{0.32\linewidth}
        \centering
        \includegraphics[width=\linewidth]{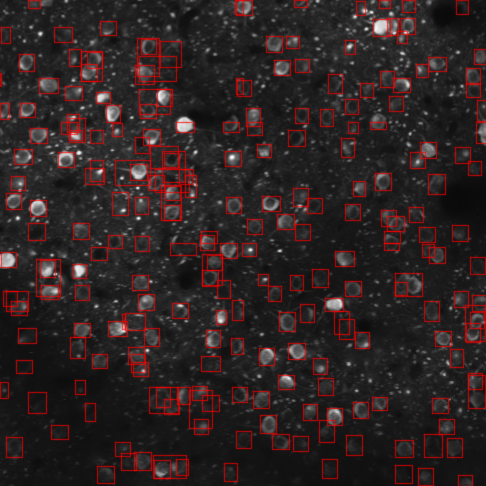}
        \caption{Recall Predictor Selector}
    \end{subfigure}

    \caption{Qualitative comparison of detection results on a representative test sample. (a) Ground truth annotations; (b) MedSAM~3 zero-shot performance without exemplar prompting; (c-f) Results obtained using different exemplar selection strategies. }
    \label{fig:medsam_grid}
\end{figure}

To complement the analysis in the main paper, we extend the evaluation of RAVP to MedSAM~3~\cite{medisam3}, testing whether the conclusions obtained with the generic SAM~3 backbone extend to a model that already incorporates medical-domain adaptation. The results in Tables \ref{tab:medsam_detection} and \ref{tab:medsam_segm} show that MedSAM~3 provides a substantially stronger zero-shot starting point, particularly under domain shift, where detection H increases from 0.204 for SAM~3 to 0.580. Nevertheless, the relative behavior of the adaptation strategies remains consistent with the main analysis: encoder-only and LoRA adaptation are generally preferable to full fine-tuning. The Library mechanism also consistently improves the MedSAM~3 zero-shot baseline, although with smaller margins than for SAM~3. The Recall Predictor performs best in-domain and on the STNeuroNet subset, whereas OOD results are less sensitive to the selection strategy, further supporting the main-paper observation that, under strong domain shift, much of the gain comes from visual prompting itself rather than from the specific retrieval policy. Qualitative results for the different exemplar selection strategies are shown in Fig.~\ref{fig:medsam_grid}, illustrating how the choice of selector affects the resulting detections.

\begin{table*}[ht]
\centering
\caption{Detection performance across ABO evaluation splits.
}
\label{tab:medsam_detection}
\setlength{\tabcolsep}{5pt}
\resizebox{\textwidth}{!}{%
\begin{tabular}{lccccccccc}
\toprule
& \multicolumn{3}{c}{\textbf{ABO in-domain}} & \multicolumn{3}{c}{\textbf{OOD}} & \multicolumn{3}{c}{\textbf{STNeuroNet subset}} \\
\cmidrule(lr){2-4}\cmidrule(lr){5-7}\cmidrule(lr){8-10}
\textbf{Method} & AP$_{50}$ & AR$_{50}$ & H & AP$_{50}$ & AR$_{50}$ & H & AP$_{50}$ & AR$_{50}$ & H \\
\midrule
MedSAM~3 (zero-shot) & 0.470 & 0.651 & 0.546 & 0.483 & 0.726 & 0.580 & 0.502 & 0.620 & 0.554 \\
\midrule
MedSAM~3 + Enc.\ FT & 0.691 & 0.836 & 0.756 & 0.641 & 0.853 & 0.732 & 0.734 & 0.815 & 0.773 \\
MedSAM~3 + LoRA & 0.651 & 0.816 & 0.724 & 0.607 & 0.833 & 0.702 & 0.727 & 0.809 & 0.766 \\
MedSAM~3 + Full FT & 0.628 & 0.806 & 0.706 & 0.585 & 0.840 & 0.690 & 0.689 & 0.784 & 0.733 \\
\midrule
MedSAM~3 + Lib. \textit{(Random)} & 0.481 & 0.630 & 0.546 & 0.506 & 0.690 & 0.584 & 0.501 & 0.600 & 0.546 \\
MedSAM~3 + Lib. \textit{(Image-Adaptive)} & 0.481 & 0.638 & 0.549 & 0.511 & 0.702 & 0.591 & 0.510 & 0.607 & 0.554 \\
MedSAM~3 + Lib. \textit{(Fixed fluorescence)} & 0.494 & 0.669 & 0.568 & 0.516 & 0.718 & 0.601 & 0.543 & 0.656 & 0.594 \\
MedSAM~3 + Lib. \textit{(Recall Predictor)} & 0.518 & 0.687 & 0.590 & 0.495 & 0.738 & 0.592 & 0.577 & 0.671 & 0.620 \\
\bottomrule
\end{tabular}
}
\end{table*}

\begin{table*}[ht]
\centering
\caption{Segmentation performance across ABO evaluation splits.
}
\label{tab:medsam_segm}
\setlength{\tabcolsep}{5pt}
\resizebox{\textwidth}{!}{%
\begin{tabular}{lccccccccc}
\toprule
& \multicolumn{3}{c}{\textbf{ABO in-domain}} & \multicolumn{3}{c}{\textbf{OOD}} & \multicolumn{3}{c}{\textbf{STNeuroNet subset}} \\
\cmidrule(lr){2-4}\cmidrule(lr){5-7}\cmidrule(lr){8-10}
\textbf{Method} & AP$_{50}$ & AR$_{50}$ & H & AP$_{50}$ & AR$_{50}$ & H & AP$_{50}$ & AR$_{50}$ & H \\
\midrule
MedSAM~3 (zero-shot) & 0.495 & 0.661 & 0.566 & 0.480 & 0.691 & 0.566 & 0.523 & 0.636 & 0.574 \\
\midrule
MedSAM~3 + Enc.\ FT & 0.703 & 0.837 & 0.764 & 0.628 & 0.834 & 0.716 & 0.728 & 0.799 & 0.762 \\
MedSAM~3 + LoRA & 0.672 & 0.828 & 0.742 & 0.606 & 0.827 & 0.699 & 0.736 & 0.805 & 0.769 \\
MedSAM~3 + Full FT & 0.650 & 0.818 & 0.724 & 0.591 & 0.833 & 0.692 & 0.691 & 0.774 & 0.730 \\
\midrule
MedSAM~3 + Lib. \textit{(Random)} & 0.525 & 0.660 & 0.585 & 0.508 & 0.676 & 0.580 & 0.543 & 0.636 & 0.586 \\
MedSAM~3 + Lib. \textit{(Image-Adaptive)} & 0.528 & 0.665 & 0.589 & 0.506 & 0.672 & 0.577 & 0.557 & 0.644 & 0.597 \\
MedSAM~3 + Lib. \textit{(Fixed fluorescence)} & 0.529 & 0.678 & 0.594 & 0.507 & 0.689 & 0.584 & 0.577 & 0.671 & 0.621 \\
MedSAM~3 + Lib. \textit{(Recall Predictor)} & 0.555 & 0.698 & 0.618 & 0.495 & 0.711 & 0.584 & 0.605 & 0.685 & 0.642 \\
\bottomrule
\end{tabular}
}
\end{table*}

\clearpage
\section{Additional Ablations}
Table~\ref{tab:supp_library_ablation} reports additional ablations of the Library mechanism in the zero-shot setting.
The first row corresponds to the original SAM~3 zero-shot baseline \cite{sam3}, without Library augmentation.
In the \textit{Black no bbox} variant, the retrieved Library example is replaced by a black image and no bounding box is provided.
The \textit{Black keep bbox} variant also replaces the Library example with a black image, but retains the bounding box associated with the retrieved example.
The \textit{Random bbox} variant keeps the retrieved Library example but replaces its bounding box with a randomly selected one.
Finally, the \textit{Recall Predictor} variant uses the complete Library mechanism, providing the retrieved example together with the corresponding bounding box selected according to the predicted recall.

These controls show that the gains obtained with the Library mechanism are not explained by the canvas layout, padding, or the mere presence of bounding boxes: the semantic content of the retrieved exemplar plays a central role.
When both the exemplar and its bounding box are replaced by black padding, performance changes only marginally with respect to zero-shot SAM~3 ($H=0.420$ vs.\ $0.407$), possibly due to the different effective image resolution induced by the composed canvas.
Conversely, retaining bounding boxes over black regions or pairing the retrieved exemplar with a random bounding box does not reproduce the gain of the complete Library mechanism ($H=0.615$), but instead degrades performance to $H=0.339$ and $H=0.239$, respectively.

\begin{table*}[htb!]
\centering
\caption{Effect of the Library mechanism in the zero-shot regime. AP$_{50}$ / AR$_{50}$ and H for bounding-box detection across ABO in-domain, OOD, and STNeuroNet subset. Best results per split and metric in \textbf{bold}.}
\label{tab:supp_library_ablation}
\setlength{\tabcolsep}{4pt}
\resizebox{\textwidth}{!}{%
\begin{tabular}{lccc|ccc|ccc}
\toprule
& \multicolumn{3}{c}{\textbf{ABO in-domain}}
& \multicolumn{3}{c}{\textbf{OOD}}
& \multicolumn{3}{c}{\textbf{STNeuroNet subset}} \\
\cmidrule(lr){2-4}
\cmidrule(lr){5-7}
\cmidrule(lr){8-10}
\textbf{Method}
& AP$_{50}$ & AR$_{50}$ & H
& AP$_{50}$ & AR$_{50}$ & H
& AP$_{50}$ & AR$_{50}$ & H \\
\midrule
SAM~3 (zero-shot)
& 0.277 & \textbf{0.767} & 0.407
& 0.117 & \textbf{0.802} & 0.204
& 0.405 & \textbf{0.736} & 0.523 \\

\quad + Lib.\ (\textit{Black no bbox})
& 0.295 & 0.730 & 0.420
& 0.169 & 0.766 & 0.276
& 0.355 & 0.699 & 0.471 \\

\quad + Lib.\ (\textit{Black keep bbox})
& 0.227 & 0.673 & 0.339
& 0.267 & 0.751 & 0.393
& 0.321 & 0.662 & 0.433 \\

\quad + Lib.\ (\textit{Random bbox})
& 0.162 & 0.461 & 0.239
& 0.245 & 0.622 & 0.352
& 0.129 & 0.445 & 0.200 \\

\quad + Lib.\ (\textit{Recall Predictor})
& \textbf{0.528} & 0.736 & \textbf{0.615}
& \textbf{0.468} & 0.786 & \textbf{0.587}
& \textbf{0.609} & 0.732 & \textbf{0.665} \\
\bottomrule
\end{tabular}
}
\end{table*}

\bibliographystyle{unsrtnat}
\bibliography{biblio}

@article{pnevmatikakis2016simultaneous,
  title={Simultaneous denoising, deconvolution, and demixing of calcium imaging data},
  author={Pnevmatikakis, Eftychios A and Soudry, Daniel and Gao, Yuanjun and Machado, Timothy A and Merel, Josh and Pfau, David and Reier, Thomas and Ahrens, Misha B and Bruno, Randy M and Jessell, Thomas M and others},
  journal={Neuron},
  volume={89},
  number={2},
  pages={285--299},
  year={2016},
  publisher={Elsevier}
}

@article{pachitariu2017suite2p,
  title={Suite2p: beyond 10,000 neurons with standard two-photon microscopy},
  author={Pachitariu, Marius and Stringer, Carsen and Dipoppa, Mario and Schr{\"o}der, Sylvia and Rossi, L Federico and Dalgleish, Henry and Carandini, Matteo and Harris, Kenneth D},
  journal={bioRxiv},
  pages={161507},
  year={2017},
  publisher={Cold Spring Harbor Laboratory}
}

@article{giovannucci2019caiman,
  title={CaImAn: An open source tool for scalable calcium imaging data analysis},
  author={Giovannucci, Andrea and Friedrich, Johannes and Gunn, Pat and Kalfon, J{\'e}r{\'e}mie and Brown, Brandon L and Koay, Sue Ann and Taxidis, Jiannis and Najafi, Farzaneh and Gauthier, Jeffrey L and Zhou, Pengcheng and others},
  journal={eLife},
  volume={8},
  pages={e38173},
  year={2019},
  publisher={eLife Sciences Publications, Ltd}
}

@article{friedrich2017fast,
  title={Fast online deconvolution of calcium imaging data},
  author={Friedrich, Johannes and Zhou, Pengcheng and Paninski, Liam},
  journal={PLoS Computational Biology},
  volume={13},
  number={3},
  pages={e1005423},
  year={2017},
  publisher={Public Library of Science San Francisco, CA USA}
}

@article{dmitrieva2024realseudo,
  title={realSEUDO for real-time calcium imaging analysis},
  author={Dmitrieva, Iuliia and Babkin, Sergey and Charles, Adam S},
  journal={arXiv preprint arXiv:2405.15701},
  year={2024}
}

@article{sita2022deep,
  title={A deep-learning approach for online cell identification and trace extraction in functional two-photon calcium imaging},
  author={Sit{\`a}, Luca and Brondi, Marco and Lagomarsino de Leon Roig, Pedro and Curreli, Sebastiano and Panniello, Mariangela and Vecchia, Dania and Fellin, Tommaso},
  journal={Nature Communications},
  volume={13},
  number={1},
  pages={1529},
  year={2022},
  publisher={Nature Publishing Group UK London}
}

@article{wu2024neuroseg,
  title={NeuroSeg-III: efficient neuron segmentation in two-photon Ca2+ imaging data using self-supervised learning},
  author={Wu, Yukun and Xu, Zhehao and Liang, Shanshan and Wang, Lukang and Wang, Meng and Jia, Hongbo and Chen, Xiaowei and Zhao, Zhikai and Liao, Xiang},
  journal={Biomedical Optics Express},
  volume={15},
  number={5},
  pages={2910--2926},
  year={2024},
  publisher={Optica Publishing Group}
}

@article{wu2022general,
  title={A General Deep Learning framework for Neuron Instance Segmentation based on Efficient UNet and Morphological Post-processing},
  author={Wu, Huaqian and Souedet, Nicolas and Jan, Caroline and Clouchoux, C{\'e}dric and Delzescaux, Thierry},
  journal={arXiv preprint arXiv:2202.08682},
  year={2022}
}

@article{sam3,
  title={Sam 3: Segment anything with concepts},
  author={Carion, Nicolas and Gustafson, Laura and Hu, Yuan-Ting and Debnath, Shoubhik and Hu, Ronghang and Suris, Didac and Ryali, Chaitanya and Alwala, Kalyan Vasudev and Khedr, Haitham and Huang, Andrew and others},
  journal={arXiv preprint arXiv:2511.16719},
  year={2025}
}

@article{medisam3,
  title={Medical SAM3: A Foundation Model for Universal Prompt-Driven Medical Image Segmentation},
  author={Jiang, Chongcong and Ding, Tianxingjian and Song, Chuhan and Tu, Jiachen and Yan, Ziyang and Shao, Yihua and Wang, Zhenyi and Shang, Yuzhang and Han, Tianyu and Tian, Yu},
  journal={arXiv preprint arXiv:2601.10880},
  year={2026}
}

@inproceedings{sam,
  title={Segment anything},
  author={Kirillov, Alexander and Mintun, Eric and Ravi, Nikhila and Mao, Hanzi and Rolland, Chloe and Gustafson, Laura and Xiao, Tete and Whitehead, Spencer and Berg, Alexander C and Lo, Wan-Yen and others},
  booktitle={Proceedings of the IEEE/CVF international conference on computer vision},
  pages={4015--4026},
  year={2023}
}

@article{sam2,
  title={Sam 2: Segment anything in images and videos},
  author={Ravi, Nikhila and Gabeur, Valentin and Hu, Yuan-Ting and Hu, Ronghang and Ryali, Chaitanya and Ma, Tengyu and Khedr, Haitham and R{\"a}dle, Roman and Rolland, Chloe and Gustafson, Laura and others},
  journal={arXiv preprint arXiv:2408.00714},
  year={2024}
}

@article{medsam,
  title={Segment anything in medical images},
  author={Ma, Jun and He, Yuting and Li, Feifei and Han, Lin and You, Chenyu and Wang, Bo},
  journal={Nature communications},
  volume={15},
  number={1},
  pages={654},
  year={2024},
  publisher={Nature Publishing Group UK London}
}

@article{lora,
  title={Lora: Low-rank adaptation of large language models.},
  author={Hu, Edward J and Shen, Yelong and Wallis, Phillip and Allen-Zhu, Zeyuan and Li, Yuanzhi and Wang, Shean and Wang, Liang and Chen, Weizhu and others},
  journal={Iclr},
  volume={1},
  number={2},
  pages={3},
  year={2022}
}

@misc{abo,
  title     = {{Allen Brain Observatory: Technical Whitepaper}},
  author    = {{Allen Institute for Brain Science}},
  year      = {2016},
  url       = {http://observatory.brain-map.org/visualcoding}
}

@article{stneuronet,
  title={Fast and robust active neuron segmentation in two-photon calcium imaging using spatiotemporal deep learning},
  author={Soltanian-Zadeh, Somayyeh and Sahingur, Kaan and Blau, Sarah and Gong, Yiyang and Farsiu, Sina},
  journal={Proceedings of the National Academy of Sciences},
  volume={116},
  number={17},
  pages={8554--8563},
  year={2019},
  publisher={National Academy of Sciences}
}

@article{suite2p,
  title={Suite2p: beyond 10,000 neurons with standard two-photon microscopy},
  author={Pachitariu, Marius and Stringer, Carsen and Schr{\"o}der, Sylvia and Dipoppa, Mario and Rossi, L Federico and Carandini, Matteo and Harris, Kenneth D},
  journal={BioRxiv},
  pages={061507},
  year={2016},
  publisher={Cold Spring Harbor Laboratory}
}

@inproceedings{painter,
  title={Images speak in images: A generalist painter for in-context visual learning},
  author={Wang, Xinlong and Wang, Wen and Cao, Yue and Shen, Chunhua and Huang, Tiejun},
  booktitle={Proceedings of the IEEE/CVF Conference on Computer Vision and Pattern Recognition},
  pages={6830--6839},
  year={2023}
}

@inproceedings{panet,
  title={Panet: Few-shot image semantic segmentation with prototype alignment},
  author={Wang, Kaixin and Liew, Jun Hao and Zou, Yingtian and Zhou, Daquan and Feng, Jiashi},
  booktitle={proceedings of the IEEE/CVF international conference on computer vision},
  pages={9197--9206},
  year={2019}
}

@inproceedings{hsnet,
  title={Hypercorrelation squeeze for few-shot segmentation},
  author={Min, Juhong and Kang, Dahyun and Cho, Minsu},
  booktitle={Proceedings of the IEEE/CVF international conference on computer vision},
  pages={6941--6952},
  year={2021}
}

@article{prototypical,
  title={Prototypical networks for few-shot learning},
  author={Snell, Jake and Swersky, Kevin and Zemel, Richard},
  journal={Advances in neural information processing systems},
  volume={30},
  year={2017}
}

@article{stosiek2003,
  title={In vivo two-photon calcium imaging of neuronal networks},
  author={Stosiek, Christoph and Garaschuk, Olga and Holthoff, Knut and Konnerth, Arthur},
  journal={Proceedings of the National Academy of Sciences},
  volume={100},
  number={12},
  pages={7319--7324},
  year={2003},
  publisher={National Academy of Sciences}
}

@article{chen2013,
  title={Ultrasensitive fluorescent proteins for imaging neuronal activity},
  author={Chen, Tsai-Wen and Wardill, Trevor J and Sun, Yi and Pulver, Stefan R and Renninger, Sabine L and Baohan, Amy and Schreiter, Eric R and Kerr, Rex A and Orger, Michael B and Jayaraman, Vivek and others},
  journal={Nature},
  volume={499},
  number={7458},
  pages={295--300},
  year={2013},
  publisher={Nature Publishing Group UK London}
}

@article{wang2023seggpt,
  title={Seggpt: Segmenting everything in context},
  author={Wang, Xinlong and Zhang, Xiaosong and Cao, Yue and Wang, Wen and Shen, Chunhua and Huang, Tiejun},
  journal={arXiv preprint arXiv:2304.03284},
  year={2023}
}

@INPROCEEDINGS{resnet,
  author={He, Kaiming and Zhang, Xiangyu and Ren, Shaoqing and Sun, Jian},
  booktitle={2016 IEEE Conference on Computer Vision and Pattern Recognition (CVPR)}, 
  title={Deep Residual Learning for Image Recognition}, 
  year={2016},
  volume={},
  number={},
  pages={770-778},
  doi={10.1109/CVPR.2016.90}}

\end{document}